\documentclass[11pt]{article}
\pdfoutput=1
\usepackage[a4paper]{geometry}
\usepackage{clic2017}
\usepackage{times}
\usepackage{hyperref}
\usepackage{url}
\usepackage{latexsym}
\usepackage{natbib}
\usepackage{amsfonts}
\usepackage{multirow}
\usepackage{booktabs}
\usepackage{graphicx}
\usepackage{xstring}
\usepackage{collcell}
\usepackage[table, dvipsnames]{xcolor}
\newcommand{\checkvalue}[1]{\IfInteger{#1}{\ifnum#1<0\ifnum#1<-15 \cellcolor{SkyBlue}\fi\else\ifnum#1>15\cellcolor{BurntOrange}\fi\fi#1\%}{#1}}
\newcolumntype{L}{>{\collectcell\checkvalue}l<{\endcollectcell}}
\newcolumntype{C}{>{\collectcell\checkvalue}c<{\endcollectcell}}
\newcolumntype{R}{>{\collectcell\checkvalue}r<{\endcollectcell}}

\title{UmBERTo-MTSA @ AcCompl-It:  \\ Improving Complexity and Acceptability Prediction \\ with Multi-task Learning on Self-Supervised Annotations}

\author{Gabriele Sarti \\
    Department of Mathematics and Geoscience, University of Trieste \\
    International School for Advanced Studies (SISSA), Trieste, Italy \\
  {\tt gsarti@sissa.it} \\}

\date{}

\newcommand\blfootnote[1]{%
  \begingroup
  \renewcommand\thefootnote{}\footnote{#1}%
  \addtocounter{footnote}{-1}%
  \endgroup
}

\begin{document}
\maketitle
\begin{abstract}
  \textbf{English.}  This work describes a self-supervised data augmentation approach used to improve learning models' performances when only a moderate amount of labeled data is available. Multiple copies of the original model are initially trained on the downstream task. Their predictions are then used to annotate a large set of unlabeled examples. Finally, multi-task training is performed on the parallel annotations of the resulting training set, and final scores are obtained by averaging annotator-specific head predictions. Neural language models are fine-tuned using this procedure in the context of the AcCompl-it shared task at EVALITA 2020, obtaining considerable improvements in prediction quality.
\end{abstract}

\begin{abstract-alt}
 \textrm{\bf{Italiano.}} Questo articolo descrive un approccio di self-supervised data augmentation utilizzabile al fine di migliorare le performance di algoritmi di apprendimento su task aventi solo una modesta quantità di dati annotati. Inizialmente, molteplici copie del modello originale vengono allenate sul task prescelto. Le loro previsioni vengono poi utilizzate per annotare grandi quantità di esempi non etichettati. In conclusione, un approccio di multi-task training viene utilizzato, con le annotazioni del dataset risultante in veste di task indipendenti, per ottenere previsioni finali come medie dei i punteggi dei singoli annotatori. Questa procedura è stata utilizzata per allenare modelli del linguaggio neurali per lo shared task AcCompl-it a EVALITA 2020, ottenendo ampi miglioramenti nella qualità predittiva.
\end{abstract-alt}

\blfootnote{Copyright $\copyright$ 2020 for this paper by its authors. Use permitted under Creative Commons License Attribution 4.0 International (CC BY 4.0).}

\section{Introduction}

In recent times, pre-trained neural language models (NLMs) have become the preferred approach for language representation learning, pushing the state-of-the-art in multiple NLP tasks~(\citet{devlin-etal-2019-bert,radford-etal-2019-language,Yang2019XLNetGA, Raffel2019ExploringTL} \textit{inter alia}). These approaches rely on a two-step training process: first, a \textit{self-supervised pre-training} is performed on large-scale corpora; then, the model undergoes a \textit{supervised fine-tuning} on downstream task labels using task-specific prediction heads. While this method was found to be effective in scenarios where a relatively large amount of labeled data are present, researchers highlighted that this is not the case in low-resource settings~\citep{Yogatama2019LearningAE}. 

Recently, \textit{pattern-exploiting training}~(PET, \citet{Schick2020ExploitingCQ,Schick2020ItsNJ} tackles the dependence of NLMs on labeled data by first reformulating tasks as cloze questions using task-related patterns and keywords, and then using language models trained on those to annotate large sets of unlabeled examples with soft labels. PET can be thought of as an offline version of \textit{knowledge distillation}~\citep{Hinton2015DistillingTK}, which is a well-established approach to transfer the knowledge across models of different size, or even between different versions of the same model as in \textit{self-training} \citep{scudder-1965-probability, yarowsky-1995-unsupervised}. While effective on classification tasks that can be easily reformulated as cloze questions, PET cannot be easily extended to regression settings since they cannot be adequately verbalized. Contemporary work by \citet{du-etal-2020-selftraining} showed how self-training and pre-training provide complementary information for natural language understanding tasks.

In this paper, I propose a simple self-supervised data augmentation approach that can be used to improve the generalization capabilities of NLMs on regression and classification tasks for modest-sized labeled corpora. In short, an ensemble of fine-tuned models is used to annotate a large corpus of unlabeled text, and new annotations are leveraged in a multi-task setting to obtain final predictions over the original test set. The method was tested on the AcCompl-it shared tasks of the EVALITA 2020 campaign~\citep{accomplit2020, Evalita2020}, where the objective was to predict respectively \textit{complexity} and \textit{acceptability} scores on a 1-7 Likert scale for each test sentence, alongside an estimation of its standard error. Results show considerable improvements over regular fine-tuning performances on COMPL and ACCEPT using the UmBERTo pre-trained model~\citep{umberto}, suggesting the validity of this approach for complexity/acceptability prediction and possibly other language processing tasks.

\section{Description of the Approach}

Let:
\begin{itemize}
    \item $\mathcal{L} = [(x_1, y_1), \dots (x_n, y_n)]$ be the initial labeled corpus containing sentence-annotation pairs $x_i \in X, y_i \in Y_x$.~\footnote{$y_i$ can be either discrete or continuous in this context.}
    \item $\mathcal{U} = [x'_1, \dots x'_m]$ be a large unlabeled corpus such that $m \gg n$
    \item $M: x_i \rightarrow \hat y_i$ be a pre-trained neural language model with a single task-specific heads, taking sentence $x_i$ as input and predicting label $y_i$ at inference time.
\end{itemize}
For some $k \in \mathbb{N}_1$, we begin by splitting $\mathcal{L}$ in $k$ equal-sized segments $\mathcal{L}_1,\dots,\mathcal{L}_k$ and fine-tune $k$ identical versions of $M$ using $k$-fold cross-validation. We call the resulting models $M^1, \dots, M^k$ ``NLMs with standard fine-tuning on the $y$ target task'', with $M^i$ being trained on the subset $\mathcal{L} - \mathcal{L}_i$ and evaluated on $\mathcal{L}_i$. Then, each sentence of $\mathcal{U}$ is passed to each model, obtaining the corpus 
\begin{equation}
    \mathcal{U}' = [(x'_1, \hat y'^1_1 \dots \hat y'^k_1),\dots,(x'_m, \hat y'^1_m \dots \hat y'^k_m)]
\end{equation}
labeled with expert annotations from fine-tuned models. Predicted values are taken instead of probability distributions after the softmax, which are typically used in the knowledge distillation literature, to keep the approach simple while making it viable in the context of regression tasks.

Now that the large corpus is annotated, a \textit{multi-task NLM} $MTM: x_i \rightarrow \dot y^1_i \dots \dot y^k_i$ is fine-tuned on $\mathcal{U}'$ by treating each annotation in the set $\hat y'^1 \dots \hat y'^k$ as a separate task, using 1-layer feed-forward neural networks as task-specific heads while performing hard parameter sharing~\citep{caruana-1997-multitask} on underlying model parameters. Intuitively, the $k$ models used to produce annotations were trained on different folds of the original corpus, and as such, they provide complementary viewpoints on the modeled phenomenon when $k$ is small.

As a final step, $MTM$ is fine-tuned on a training portion of $\mathcal{L}$, using as prediction scores $f(\dot y^1_i \dots \dot y^k_i)$, where $f$ is a task and context-dependent aggregation function. For example, in the case of a classification task, one can select the majority vote from the ensemble of model heads as the final prediction, while in a regression setting this can be done by averaging scores across heads. Once fine-tuned, the model can be tested on the test portion of $\mathcal{L}$ using the same $f$ as the aggregator. I refer to this approach as \textit{Multi-Task Self-Annotation (MTSA)} in the following sections.

\section{Experimental Evaluation}

For the experimental evaluation part:

\begin{itemize}
    \item The ACCEPT and COMPL training corpora, containing respectively 1339 and 2012 sentences labeled with average scores and standard error across annotators, were used as labeled datasets $\mathcal{L}_A, \mathcal{L}_C$. The two tasks were learned separately, following the same approach described in the previous section.
    \item A set of multiple Italian treebanks including train, dev, and test sets of the Italian Stanford Dependency Treebank~\citep{bosco-etal-2013-converting}, the Turin University Parallel Treebank~\citep{sanguinetti-etal-2015-partut}, PoSTWITA-UD~\citep{sanguinetti-etal-2018-postwita} and the Venice Italian Treebank~\citep{delmonte2007vit} was used as unlabeled corpus $\mathcal{U}$. The final corpus contains 37,344 unlabeled sentences and spans multiple textual genres.
    \item The UmBERTo model~\citep{umberto} available through the HuggingFace's Transformers framework~\citep{Wolf2019HuggingFacesTS} was used both for fine-tuning $M^{1\dots k}$ during the annotation part and for fine-tuning $MTM$. The model is based on the RoBERTa architecture~\citep{Liu2019RoBERTaAR} and was pre-trained on the Italian portion of the OSCAR CommonCrawl corpus~\citep{ortiz-suarez-etal-2020-monolingual}, containing roughly 210M sentences and over 11B tokens.
\end{itemize}

\begin{table}
\begin{center}
\begin{tabular}{lcc}
\toprule
\textbf{Model} & \textbf{Score} ($\rho$) & \textbf{Error} ($\rho$) \\
\midrule
UmBERTo surprisal     &  -0.36  & 0.17  \\
Length (\# of tokens) &  -0.39  & 0.17  \\
Length (characters)   &  -0.39  & 0.21  \\
UmBERTo fine-tuned    &  0.90   & 0.50  \\
UmBERTo-STSA          &  \textbf{0.91}   & 0.53  \\
UmBERTo-MTSA          &  \textbf{0.91}   & \textbf{0.54}  \\
\midrule
UmBERTo surprisal     &  0.49  & 0.28  \\
Length (\# of tokens) &  0.55  & 0.36  \\
Length (characters)   &  0.60  & 0.39  \\
UmBERTo fine-tuned    &  0.84  & 0.54  \\
UmBERTo-STSA          &  0.87  & 0.62  \\
UmBERTo-MTSA    &  \textbf{0.88}  & \textbf{0.63}  \\
\bottomrule
\end{tabular}
\caption{Spearman's correlation scores on the ACCEPT (top) and COMPL (bottom) subtasks' training portions. Models are evaluated using 5-fold cross-validation. All scores have $p < 0.001$}\label{tab:tests}
\end{center}
\end{table}

Since both tasks involve predicting both averaged scores and the original standard error across participants, the approach presented in the previous section was adapted to account for multi-task learning of scores and errors from the beginning, with each model $M^i$ producing both a predicted score $\hat y'^i$ and a predicted error $\hat \epsilon'^i$ for the annotation step. The $k$ parameter was set to 5 to prevent excessive overlapping of training data across models, with the final multi-task model $MTM: x_i \rightarrow \dot y^1_i\dots y^5_i, \epsilon^1_i\dots \epsilon^5_i$ returning prediction for scores and errors for all the five sets of fine-tuned model annotations.

Models $M^{1\dots k}$ were trained for a maximum of 15 epochs on the labeled training sets using early stopping (5 patience steps, 20 evaluation steps using a 10\% slice as dev set), learning rate $\lambda = 1e^{-5}$, batch size $b = 32$ and embedding dropout $\delta = 0.1$. The model's base variant was used, having a hidden size $|h| = 768$, and a maximum sequence length of 128. Notably, the representations at the last layer of the UmBERTo model were averaged to obtain a sentence-level representation instead of using the [CLS] token. During the training on the whole unlabeled corpus, the evaluation steps were increased to 100 to balance evaluation time with the corpus's increased size.

\section{Results}

Table~\ref{tab:tests} presents methods for which the correlation between values and complexity scores was tested on the training portion of the ACCEPT and COMPL tasks with 5-fold cross validation, leading to the selection of MTSA as the top-performing approach:

\begin{itemize}
    \item \textbf{UmBERTo surprisal}: Sentence-level surprisal estimates are produced using the pre-trained model without fine-tuning as:
    \begin{equation}
    P(x) = \prod_{i=1}^m P(w_i \ | w_{1:i-1}, w_{i+1:m})
    \end{equation}
    \item \textbf{Length (\# of tokens)}: Length of the sentence in number of tokens
    \item \textbf{Length (characters)}: Length of the sentence in number of characters (including whitespaces)
    \item \textbf{UmBERTo fine-tuned}: Predictions produced by Umberto with standard fine-tuning on complexity corpus annotations.
    \item \textbf{UmBERTo-STSA}: A variant of the MTSA approach where instead of performing multi-task learning over model annotations on $\mathcal{U}$, we average them in a single score, and the model is trained on it with single-task fine-tuning.
    \item \textbf{UmBERTo-MTSA}: The approach presented in this work.
\end{itemize}

\begin{table}
\begin{center}
\begin{tabular}{lcc}
\toprule
\textbf{Model} & \textbf{Score} ($\rho$) & \textbf{Error} ($\rho$) \\
\midrule
SVM 2-gram baseline & 0.30 & 0.35 \\
UmBERTo-MTSA          & \textbf{0.88} & \textbf{0.52} \\
\midrule
SVM length baseline & 0.50 & 0.33 \\
UmBERTo-MTSA          & \textbf{0.83} & \textbf{0.51} \\
\bottomrule
\end{tabular}
\caption{Correlation scores with gold labels on the ACCEPT (top) and COMPL (bottom) subtasks' test portions. All scores have $p < 0.001$.}\label{tab:task-scores}
\end{center}
\end{table}

From Table~\ref{tab:tests}, it can be observed that, although length alone is already correlated with acceptability complexity scores, UmBERTo can leverage additional information from its representation to produce much stronger predictions. Interestingly, both the STSA and MTSA self-annotation approaches consistently outperform regular fine-tuning, especially for what concerns standard error scores. This fact suggests that self-annotation leads to better generalization capabilities in the model over downstream tasks when relatively few annotations are available. While the contribution of multi-task learning is modest, the MTSA approach may prove especially beneficial when training models $M^{1\dots k}$ on scores produced by different annotators instead of using different folds of the same corpus, as in this case. In both cases, predicted surprisal scores act as poor predictors for downstream tasks. It should also be noted that length appears to be negatively correlated to acceptability scores (i.e. longer sentences are generally less acceptable), while the relation is positive in the case of complexity (i.e. longer sentences are generally more complex).

Table~\ref{tab:task-scores} reports the scores obtained by MTSA over the test sets for the ACCEPT and the COMPL shared tasks. The organizers' baseline scores correspond to the correlation among gold labels and acceptability and complexity predictions produced by an SVM model trained on 1-grams and bigrams of sentences and an SVM trained on sentence length, respectively. The MTSA approach achieved the first rank in both tasks, with considerable improvements over baseline scores.

\section{Error Analysis}

Finally, some error analysis is performed to gain additional insights on which factors influence the predictability of complexity and acceptability judgments. The Profiling-UD tool by~\citet{profilingud-brunato-2020} is used to produce linguistic annotations on test sentences for both tasks. Given an input sentence, Profiling-UD produces roughly $\sim100$ numeric scores representing different phenomena and properties at different language levels.\footnote{A description of produced annotations is omitted for brevity. Refer to \citet{profilingud-brunato-2020} for additional details.} I then correlate the value of all features with $y_\epsilon$ and $\epsilon_\epsilon$, representing the mean absolute error between true and predicted values for scores and standard errors, respectively. Table~\ref{tab:error-analysis} presents the results of the error analysis.

\begin{table}
\begin{small}
\begin{center}
\begin{tabular}{l@{\qquad}CC|CC}
\toprule
& \multicolumn{2}{c}{\textbf{Acceptability}} &  \multicolumn{2}{c}{\textbf{Complexity}} \\
\cmidrule(lr){2-3}
\cmidrule(lr){4-5}
& $\rho(y_\epsilon)$ & $\rho(\epsilon_\epsilon)$ & $\rho(y_\epsilon)$ & $\rho(\epsilon_\epsilon)$  \\
\midrule
avg. score $(y)$        &  -25 &  10  &  41  &  -2  \\
std. error $(\epsilon)$ &  12  &  2   &  23 &  27 \\
\midrule
upos\_dist\_PROPN       &  19  &  -3  &  4  &  6  \\
dep\_dist\_nmod         &  19  &  -8  &  4  &  1  \\
avg\_max\_depth         &  16  &  -3  &  7  &  -7  \\
n\_prep\_chains         &  16  &  -8  &  4  &  -2  \\
prep\_chain\_len        &  16  &  -6  &  9  &  -4  \\
upos\_dist\_PRON        &  1  &  20  &  8  &  9  \\
dep\_dist\_root         &  -9  &  18  &  -4  &  23  \\
dep\_dist\_punct        &  -9  &  17  &  1  &  -3  \\
aux\_mood\_dist\_Imp    &  7  &  6  &  17  &  7  \\ 
n\_tokens               &  9  &  -13  &  5  &  -18  \\
avg\_links\_len         &  -3  &  1  &  -6  &  -17  \\
max\_links\_len         &  -1  &  -9  & -1  &  -16  \\
\bottomrule
\end{tabular}
\end{center}
\end{small}
\caption{Pearson's correlation scores between prediction errors and various linguistic features. \textcolor{BurntOrange}{Orange} and \textcolor{Cyan}{cyan} cells contain respectively positive and negative scores for which $p < 0.001$.}
\label{tab:error-analysis}
\end{table}

Strongly correlated values in Table~\ref{tab:error-analysis} correspond to features that highly influence, either positively or negatively, the prediction capabilities of the MTSA model. Extreme task scores (avg. score), denoting either not very acceptable or highly complex sentences, are less predictable than their average counterparts by MTSA. Sentences for whose the standard deviation of scores is high across participants appear to be less predictable in the context of complexity scores, while this does not affect acceptability predictions. 

Concerning acceptability, I found a significant correlation between acceptability prediction errors and the presence of multilevel syntactic structures, (\textit{avg\_max\_depth}) multiple long prepositional chains (\textit{n\_prep\_chains}, \textit{prep\_chain\_len}) and nominal modifiers (\textit{dep\_dist\_nmod}). From the complexity viewpoint, instead, the presence of inflectional morphology related to the imperfect tense in auxiliaries (\textit{aux\_mood\_dist\_Imp}) was the only property related to higher prediction errors. However, high token counts (\textit{n\_tokens}) and long dependency links (\textit{avg\_links\_len}, \textit{max\_links\_len}) were shown to make the variability in complexity scores more predictable.

Overall, results suggest that incorporating syntactic information during the model's training process may further improve complexity and acceptability models.

\section{Discussion and Conclusion}

This work introduced a simple and effective data augmentation approach improving the fine-tuning performances of NLMs when only a modest amount of labeled data is available. The approach was first formalized and then empirically tested on the ACCEPT and COMPL shared tasks of the EVALITA 2020 campaign. Strong performances were reported for both acceptability and complexity prediction using a multi-task self-training approach, obtaining the top position in both subtasks. Finally, an error analysis highlighted the unpredictability of extreme scores and sentences having complex syntactic structures.

The suggested approach, although computationally refined and well-performing, is lacking in terms of complexity-driven biases that may prove useful in the context of complexity and acceptability prediction. A possible extension of this work may include a complementary syntactic task (e.g., biaffine parsing, as in \citet{Glavas2020IsSS}) during multi-task learning to see if forcing syntactically-competent representations in the top layers may prove beneficial in the context of syntax-heavy tasks like complexity and acceptability prediction. Moreover, it would be interesting to evaluate multi-task learning performances with complexity and acceptability parallel annotations given the conceptual similarity between the two tasks and estimate the effectiveness of a feed-forward network as the final aggregator $f$ in the MTSA paradigm instead of merely averaging predictions. Finally, \citet{du-etal-2020-selftraining} findings suggest that using an unsupervised in-domain filtering approach may further improve the self-training procedure when large unlabeled corpora are available.

\section*{Acknowledgments}

The author was supported by a scholarship for Data Science and Scientific Computing students from the International School of Advanced Studies (SISSA).

\bibliography{evalita}

\begin{thebibliography}{24}
\expandafter\ifx\csname natexlab\endcsname\relax\def\natexlab#1{#1}\fi

\bibitem[{Basile et~al.(2020)Basile, Croce, Di~Maro, and Passaro}]{Evalita2020}
Valerio Basile, Danilo Croce, Maria Di~Maro, and Lucia~C. Passaro. 2020.
\newblock {EVALITA} 2020: Overview of the 7th evaluation campaign of natural
  language processing and speech tools for italian.
\newblock In \emph{Proceedings of Seventh Evaluation Campaign of Natural
  Language Processing and Speech Tools for Italian. Final Workshop (EVALITA
  2020)}, Online. CEUR.org.

\bibitem[{Bosco et~al.(2013)Bosco, Montemagni, and
  Simi}]{bosco-etal-2013-converting}
Cristina Bosco, Simonetta Montemagni, and Maria Simi. 2013.
\newblock \href {https://www.aclweb.org/anthology/W13-2308} {Converting
  {I}talian treebanks: Towards an {I}talian {S}tanford dependency treebank}.
\newblock In \emph{Proceedings of the 7th Linguistic Annotation Workshop and
  Interoperability with Discourse}, pages 61--69, Sofia, Bulgaria. Association
  for Computational Linguistics.

\bibitem[{Brunato et~al.(2020{\natexlab{a}})Brunato, Cimino, Dell'Orletta,
  Venturi, and Montemagni}]{profilingud-brunato-2020}
Dominique Brunato, Andrea Cimino, Felice Dell'Orletta, Giulia Venturi, and
  Simonetta Montemagni. 2020{\natexlab{a}}.
\newblock \href {https://www.aclweb.org/anthology/2020.lrec-1.883}
  {{Profiling-UD}: a tool for linguistic profiling of texts}.
\newblock In \emph{Proceedings of The 12th Language Resources and Evaluation
  Conference}, pages 7147--7153, Marseille, France. European Language Resources
  Association.

\bibitem[{Brunato et~al.(2020{\natexlab{b}})Brunato, Cristiano, Dell’Orletta,
  Montemagni, Venturi, and Zamparelli}]{accomplit2020}
Dominique Brunato, Chesi Cristiano, Felice Dell’Orletta, Simonetta
  Montemagni, Giulia Venturi, and Roberto Zamparelli. 2020{\natexlab{b}}.
\newblock {AcCompl-it @ EVALITA2020}: Overview of the acceptability \&
  complexity evaluation task for italian.
\newblock In \emph{Proceedings of Seventh Evaluation Campaign of Natural
  Language Processing and Speech Tools for Italian. Final Workshop (EVALITA
  2020)}, Online. CEUR.org.

\bibitem[{Caruana(1997)}]{caruana-1997-multitask}
Rich Caruana. 1997.
\newblock \href
  {https://www.cs.utexas.edu/~kuipers/readings/Caruana-mlj-97.pdf} {Multitask
  learning}.
\newblock \emph{Machine Learning}, 28:41--75.

\bibitem[{Delmonte et~al.(2007)Delmonte, Bristot, and
  Tonelli}]{delmonte2007vit}
Rodolfo Delmonte, Antonella Bristot, and Sara Tonelli. 2007.
\newblock {VIT}--venice italian treebank: syntactic and quantitative features.

\bibitem[{Devlin et~al.(2019)Devlin, Chang, Lee, and
  Toutanova}]{devlin-etal-2019-bert}
Jacob Devlin, Ming-Wei Chang, Kenton Lee, and Kristina Toutanova. 2019.
\newblock \href {https://doi.org/10.18653/v1/N19-1423} {{BERT}: Pre-training of
  deep bidirectional transformers for language understanding}.
\newblock In \emph{Proceedings of the 2019 Conference of the North {A}merican
  Chapter of the Association for Computational Linguistics: Human Language
  Technologies, Volume 1 (Long and Short Papers)}, pages 4171--4186,
  Minneapolis, Minnesota. Association for Computational Linguistics.

\bibitem[{Du et~al.(2020)Du, Grave, Gunel, Chaudhary, Çelebi, Auli, Stoyanov,
  and Conneau}]{du-etal-2020-selftraining}
Jingfei Du, E.~Grave, Beliz Gunel, Vishrav Chaudhary, Onur Çelebi, M.~Auli,
  Ves Stoyanov, and Alexis Conneau. 2020.
\newblock Self-training improves pre-training for natural language
  understanding.
\newblock \emph{ArXiv}, abs/2010.02194.

\bibitem[{Francia et~al.(2020)Francia, Parisi, and Paolo}]{umberto}
Simone Francia, Loreto Parisi, and Magnani Paolo. 2020.
\newblock \href {https://github.com/musixmatchresearch/umberto} {{UmBERTo}: an
  italian language model trained with whole word maskings}.

\bibitem[{Glavas and Vulic(2020)}]{Glavas2020IsSS}
Goran Glavas and Ivan Vulic. 2020.
\newblock Is supervised syntactic parsing beneficial for language
  understanding? an empirical investigation.
\newblock \emph{ArXiv}, abs/2008.06788.

\bibitem[{Hinton et~al.(2015)Hinton, Vinyals, and
  Dean}]{Hinton2015DistillingTK}
Geoffrey~E. Hinton, Oriol Vinyals, and J.~Dean. 2015.
\newblock Distilling the knowledge in a neural network.
\newblock \emph{ArXiv}, abs/1503.02531.

\bibitem[{Liu et~al.(2019)Liu, Ott, Goyal, Du, Joshi, Chen, Levy, Lewis,
  Zettlemoyer, and Stoyanov}]{Liu2019RoBERTaAR}
Y.~Liu, Myle Ott, Naman Goyal, Jingfei Du, Mandar Joshi, Danqi Chen, Omer Levy,
  M.~Lewis, L.~Zettlemoyer, and V.~Stoyanov. 2019.
\newblock {RoBERTa}: A robustly optimized bert pretraining approach.
\newblock \emph{ArXiv}, abs/1907.11692.

\bibitem[{Ortiz~Su{\'a}rez et~al.(2020)Ortiz~Su{\'a}rez, Romary, and
  Sagot}]{ortiz-suarez-etal-2020-monolingual}
Pedro~Javier Ortiz~Su{\'a}rez, Laurent Romary, and Beno{\^\i}t Sagot. 2020.
\newblock \href {https://www.aclweb.org/anthology/2020.acl-main.156} {A
  monolingual approach to contextualized word embeddings for mid-resource
  languages}.
\newblock In \emph{Proceedings of the 58th Annual Meeting of the Association
  for Computational Linguistics}, pages 1703--1714, Online. Association for
  Computational Linguistics.

\bibitem[{Radford et~al.(2019)Radford, Wu, Child, Luan, Amodei, and
  Sutskever}]{radford-etal-2019-language}
A.~Radford, Jeffrey Wu, R.~Child, David Luan, Dario Amodei, and Ilya Sutskever.
  2019.
\newblock Language models are unsupervised multitask learners.
\newblock OpenAI.

\bibitem[{Raffel et~al.(2019)Raffel, Shazeer, Roberts, Lee, Narang, Matena,
  Zhou, Li, and Liu}]{Raffel2019ExploringTL}
Colin Raffel, Noam Shazeer, Adam Roberts, Katherine Lee, Sharan Narang, Michael
  Matena, Yanqi Zhou, W.~Li, and P.~Liu. 2019.
\newblock Exploring the limits of transfer learning with a unified text-to-text
  transformer.
\newblock \emph{ArXiv}, abs/1910.10683.

\bibitem[{Sanguinetti and Bosco(2015)}]{sanguinetti-etal-2015-partut}
Manuela Sanguinetti and Cristina Bosco. 2015.
\newblock \href {https://doi.org/10.1007/978-3-319-14206-7\_3}
  {\emph{{P}art{TUT}: The Turin University Parallel Treebank}}, pages 51--69.
  Springer International Publishing, Cham.

\bibitem[{Sanguinetti et~al.(2018)Sanguinetti, Bosco, Lavelli, Mazzei,
  Antonelli, and Tamburini}]{sanguinetti-etal-2018-postwita}
Manuela Sanguinetti, Cristina Bosco, Alberto Lavelli, Alessandro Mazzei, Oronzo
  Antonelli, and Fabio Tamburini. 2018.
\newblock \href {https://www.aclweb.org/anthology/L18-1279} {{P}o{STWITA}-{UD}:
  an {I}talian {T}witter treebank in {U}niversal {D}ependencies}.
\newblock In \emph{Proceedings of the Eleventh International Conference on
  Language Resources and Evaluation ({LREC} 2018)}, Miyazaki, Japan. European
  Language Resources Association (ELRA).

\bibitem[{Schick and Schutze(2020{\natexlab{a}})}]{Schick2020ExploitingCQ}
Timo Schick and Hinrich Schutze. 2020{\natexlab{a}}.
\newblock Exploiting cloze questions for few-shot text classification and
  natural language inference.
\newblock \emph{ArXiv}, abs/2001.07676.

\bibitem[{Schick and Schutze(2020{\natexlab{b}})}]{Schick2020ItsNJ}
Timo Schick and Hinrich Schutze. 2020{\natexlab{b}}.
\newblock It's not just size that matters: Small language models are also
  few-shot learners.
\newblock \emph{ArXiv}, abs/2009.07118.

\bibitem[{Scudder(1965)}]{scudder-1965-probability}
H~Scudder. 1965.
\newblock Probability of error of some adaptive pattern-recognition machines.
\newblock \emph{IEEE Transactions on Information Theory}, 11(3):363--371.

\bibitem[{Wolf et~al.(2019)Wolf, Debut, Sanh, Chaumond, Delangue, Moi, Cistac,
  Rault, Louf, Funtowicz, and Brew}]{Wolf2019HuggingFacesTS}
Thomas Wolf, Lysandre Debut, Victor Sanh, Julien Chaumond, Clement Delangue,
  Anthony Moi, Pierric Cistac, Tim Rault, R'emi Louf, Morgan Funtowicz, and
  Jamie Brew. 2019.
\newblock Huggingface's transformers: State-of-the-art natural language
  processing.
\newblock \emph{ArXiv}, abs/1910.03771.

\bibitem[{Yang et~al.(2019)Yang, Dai, Yang, Carbonell, Salakhutdinov, and
  Le}]{Yang2019XLNetGA}
Z.~Yang, Zihang Dai, Y.~Yang, J.~Carbonell, R.~Salakhutdinov, and Quoc~V. Le.
  2019.
\newblock {XLNet}: Generalized autoregressive pretraining for language
  understanding.
\newblock In \emph{NeurIPS}.

\bibitem[{Yarowsky(1995)}]{yarowsky-1995-unsupervised}
David Yarowsky. 1995.
\newblock \href {https://doi.org/10.3115/981658.981684} {Unsupervised word
  sense disambiguation rivaling supervised methods}.
\newblock In \emph{33rd Annual Meeting of the Association for Computational
  Linguistics}, pages 189--196, Cambridge, Massachusetts, USA. Association for
  Computational Linguistics.

\bibitem[{Yogatama et~al.(2019)Yogatama, de~Masson~d'Autume, Connor,
  Kocisk{\'y}, Chrzanowski, Kong, Lazaridou, Ling, Yu, Dyer, and
  Blunsom}]{Yogatama2019LearningAE}
Dani Yogatama, Cyprien de~Masson~d'Autume, J.~Connor, Tom{\'a}s Kocisk{\'y},
  M.~Chrzanowski, Lingpeng Kong, A.~Lazaridou, W.~Ling, L.~Yu, Chris Dyer, and
  P.~Blunsom. 2019.
\newblock Learning and evaluating general linguistic intelligence.
\newblock \emph{ArXiv}, abs/1901.11373.

\end{thebibliography}

\bibliographystyle{acl}

\end{document}